\documentclass[12pt,a4paper]{article}
\usepackage[a4paper,top=2.75cm,bottom=1.5cm,left=3.0cm,right=3.0cm,includefoot]{geometry}
\usepackage{microtype}
\usepackage{authblk}
\usepackage[utf8]{inputenc}

\usepackage[colorlinks=true,linkcolor=blue,urlcolor=blue,citecolor=blue]{hyperref}
\usepackage[hyperpageref]{backref}
\usepackage[pdftex]{graphicx}
\usepackage[round]{natbib}
\usepackage[hypcap=false]{caption}

\usepackage{amsmath}
\usepackage{amssymb}

\usepackage{tikz}

\renewcommand*{\backref}[1]{}
\renewcommand*{\backrefalt}[4]{({
    \ifcase #1 Not cited.%
          \or p.~#2%
          \else p.~#2%
    \fi%
    }).}

\newcounter{nbdrafts}
\makeatletter
\newcommand{\checknbdrafts}{
\ifnum \thenbdrafts > 0
\@latex@warning@no@line{*WARNING* The document contains \thenbdrafts \space draft note(s)}
\fi}

\makeatother

\newcommand{\nnC}{\mathcal C}
\newcommand{\nnB}{\mathcal B}
\newcommand{\nnR}{\mathcal R}
\newcommand{\nnM}{\mathcal M}
\newcommand{\nnI}{\mathcal I}
\newcommand{\nnt}{\,\rhd\,}

\title{Predicting the dynamics of 2d objects \\ with a deep residual
  network}

\author{Fran\c cois Fleuret\thanks{\tt francois.fleuret@idiap.ch}}

\affil{Computer Vision and Learning Group \\ Idiap Research Institute}

\begin{document}

\maketitle

\begin{abstract}

\setlength{\parindent}{0cm}
\setlength{\parskip}{12pt}

We investigate how a residual network can learn to predict the
dynamics of interacting shapes purely as an image-to-image regression
task.

With a simple 2d physics simulator, we generate short sequences
composed of rectangles put in motion by applying a pulling force at a
point picked at random. The network is trained with a quadratic loss
to predict the image of the resulting configuration, given the image
of the starting configuration and an image indicating the point of
grasping.

Experiments show that the network learns to predict accurately the
resulting image, which implies in particular that (1) it segments
rectangles as distinct components, (2) it infers which one contains
the grasping point, (3) it models properly the dynamic of a single
rectangle, including the torque, (4) it detects and handles collisions
to some extent, and (5) it re-synthesizes properly the entire scene
with displaced rectangles.


\end{abstract}


\section{Problem definition}

We implemented a simple 2d physics simulator to generate short
sequences of interacting shapes. The simulation is quite crude but
still includes an elastic collision model, a proper torque model, and
(strong) fluid frictions.

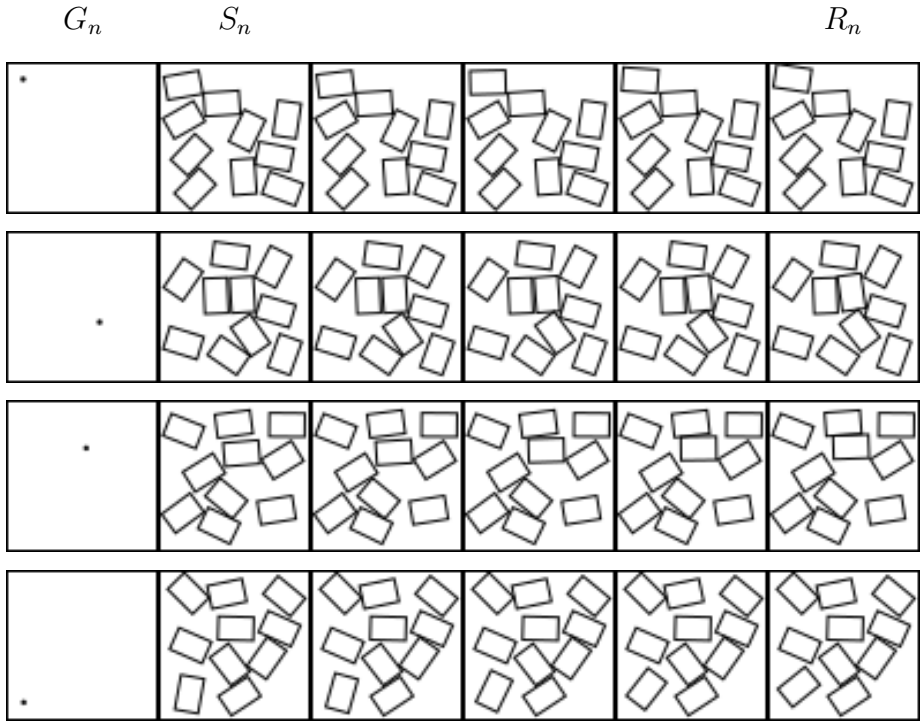
\begin{figure}[ht!]
\begin{center}
\begin{tikzpicture}
\node (n1) at (1, 0) {${{G_n}}$};
\node (n1) at (3, 0) {${{S_n}}$};
\node (n1) at (11, 0) {${{R_n}}$};
\end{tikzpicture}

\vspace*{0.5em}

\includegraphics[height=2cm]{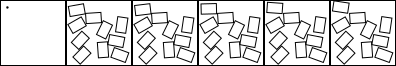}

\vspace*{0.5em}

\includegraphics[height=2cm]{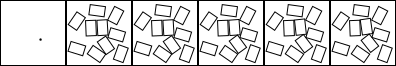}

\vspace*{0.5em}

\includegraphics[height=2cm]{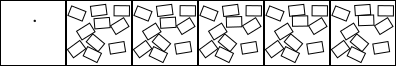}

\vspace*{0.5em}

\includegraphics[height=2cm]{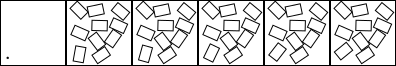}

\end{center}
\caption{Each row corresponds to one sequence of our data-set. It is
  composed of six gray-scale images of size $64 \times 64$: the
  ``grasping point image'', followed by five frames. In each sequence,
  the rectangle originally containing the grasping point is pulled
  upward and moves accordingly. It may collide with and push other
  rectangles. We show several frames of each sequence for clarity
  here, but use only the two leftmost images $(S_n, G_n)$ and the
  rightmost one $R_n$ from each sequence in our experiments, in which we
  try to predict the latter from the former.\label{fig:sequences}}
\end{figure}

As illustrated with a few examples on Figure~\ref{fig:sequences}, each
sequence is composed of gray-scale images of resolution $64 \times
64$, and is created as follows: We dispatch $10$ rectangles of fixed
size at random in the unit square, so that they do not overlap. Then
we pick at random a point uniformly in the union of the rectangle
interiors, and we apply there a constant force pulling upward for a
constant time delay. This moves the grasped rectangle upward and may
induce collisions with other rectangles, and make them move. The
borders of the square area are impenetrable, hence rectangles grabbed
near the top may have their motion constrained accordingly.

While the grasping point location is randomized for every sequence,
the characteristics of the force and its duration are common to all
the sequences.

As illustrated on Figure~\ref{fig:sequences}, from each generated
sequence we produce three images: ${{G_n}}, {{S_n}}, {{R_n}}$ which
correspond, respectively, to the grasping point image (all white, with
a dot at the location of the grasp, as shown in the leftmost column of
Figure~\ref{fig:sequences}), the starting configuration, which is the
first image of the sequence, and the resulting configuration, which is
the last image of the sequence.


\section{Network and training}

We train a residual network~\citep{resnet2015} with $18$ layer and
$16$ channels to predict ${{R_n}}$, given ${{G_n}}$ and ${{S_n}}$ as
input.

To ease the reading of long compositions of mappings, given two
mappings $f$ and $g$, let $f \nnt g$ stand for $g \circ f$.

\subsection{Structure of the network}

Our network follows the classical structure of the residual networks,
and chains several identical modules of two convolutional layers. We
define
\begin{itemize}
\item $\nnC^{f}_{c, d}$ a standard convolution layer~\citep{lecun1998}
  with filters of size $f \times f$, padding of $(f-1)/2$ to maintain the
  map size, $c$ channels as input and $d$ channels as output,
\item $\nnR$ a ReLU rectifier layer~\citep{rectifier2011},
\item $\nnB$ a batch-normalization layer~\citep{batchnormalization2015},
\item $\nnI$ the identity layer, and
\item $\nnM = \left( \nnC^{f}_{{{q}}, {{q}}} \nnt \nnB \nnt \nnR \nnt
  \nnC^{f}_{{{q}}, {{q}}} + \nnI \right) \nnt \nnB \nnt \nnR$ a two-layer
  resnet module~\citep{resnet2015} with the second batch normalization
  and non-linearity applied after summing the identity.
\end{itemize}

The structure of the full network is
\begin{equation}
\Psi = \nnC^{f}_{{{2}}, {{q}}} \nnt \nnB \nnt \nnR \nnt \underbrace{\,\nnM \nnt \dots \nnt \nnM\,}_{\times {{D}}} \nnt \nnC^{f}_{{{q}}, {{1}}}
\end{equation}
with convolution filters of size $f \times f$ with $f = 5$, ${{q}}=16$
channels in the internal encoding, and ${{D}}=8$ resnet modules, each
with two layers. It has a total of $104,417$ parameters, which is
roughly $f^2 \times q^2 \times 2D$.

\subsection{Loss, initialization and training}

We minimize the quadratic loss between the predicted and target
training images,
\begin{equation}
  L = \sum_n \| {\Psi({{S_n}}, {{G_n}})} - {{R_n}} \|_2^2
\end{equation}
and train with $32,768$ samples. We use a standard stochastic gradient
descent, randomizing the training set ordering for every epoch, using
mini-batches of size $128$, and a constant learning rate of $0.1$.

The initialization of the weights is the standard Torch rule, which
for the convolution layers is a centered uniform distribution of width
twice the inverse of the square root of the number of weights
(i.e. total number of filter coefficients), and for the batch
normalization picks the target standard deviation uniformly in $[0,1]$
and sets the target mean to zero.

We did not tune the network structure, all the results obtained here
are with the first attempt. A run with half the channels
(i.e. ${{q}}=8$) shows that it degrades noticeably the performance.


\section{Results}

\begin{figure}[ht!]
  \center
  {\includegraphics[scale=1.25,clip,trim=35 0 30 0]{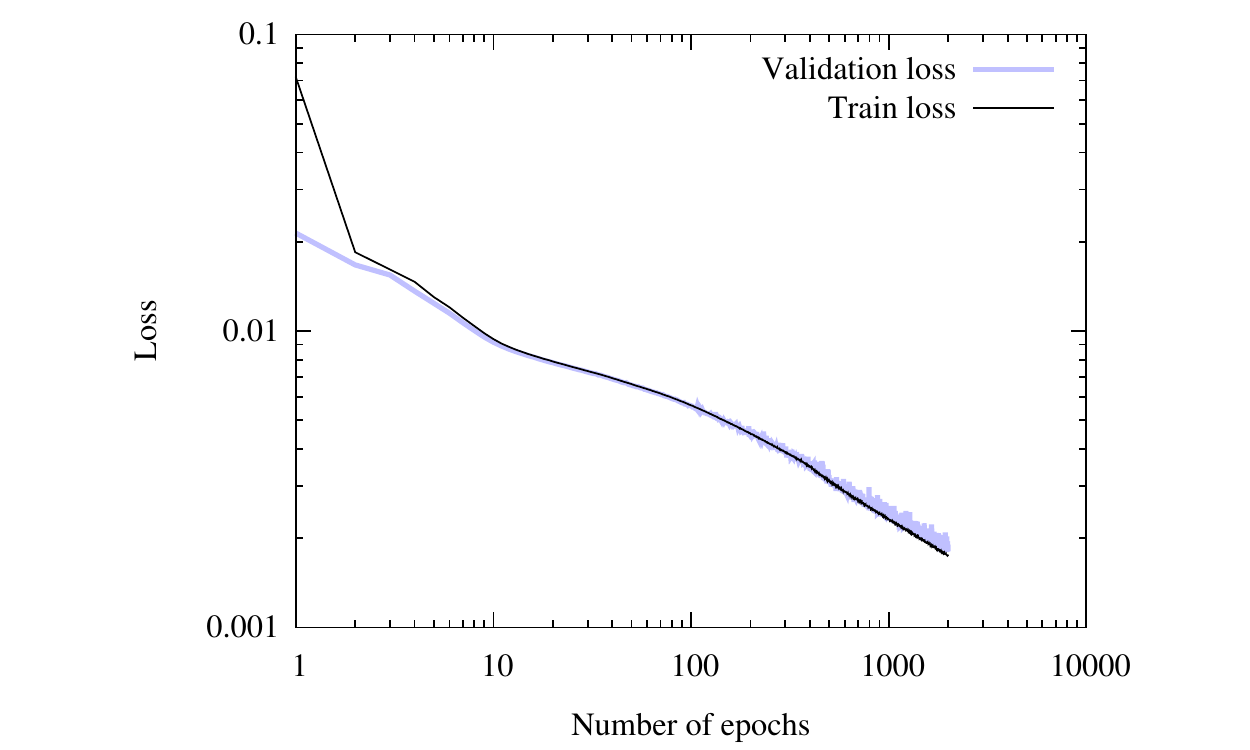}}
  \caption{Train and validation losses during training.\label{fig:losses}}
\end{figure}

We implemented the simulator in C++ and the network processing and
performance evaluation in the Torch framework~\citep{torch2011}. The
network implementation is given in
appendix~\ref{sec:torch-netw-struct}.

The source code of the simulator and the residual network to replicate
the experiments is available under the GPL-3.0 open-source license at

\vspace*{0.75em}

\centerline{\url{https://gitlab.idiap.ch/francois.fleuret/dyncnn/}}

As shown on Figure~\ref{fig:losses}, the loss decreases regularly,
with no over-fitting. It is still going down after $2,000$ epochs,
which takes slightly less than $30$ hours on a NVIDIA GTX 1080 graphic
card, using cuda toolkit 8.0, and cudnn 5.1.

\subsection{Prediction}\label{sec:prediction}

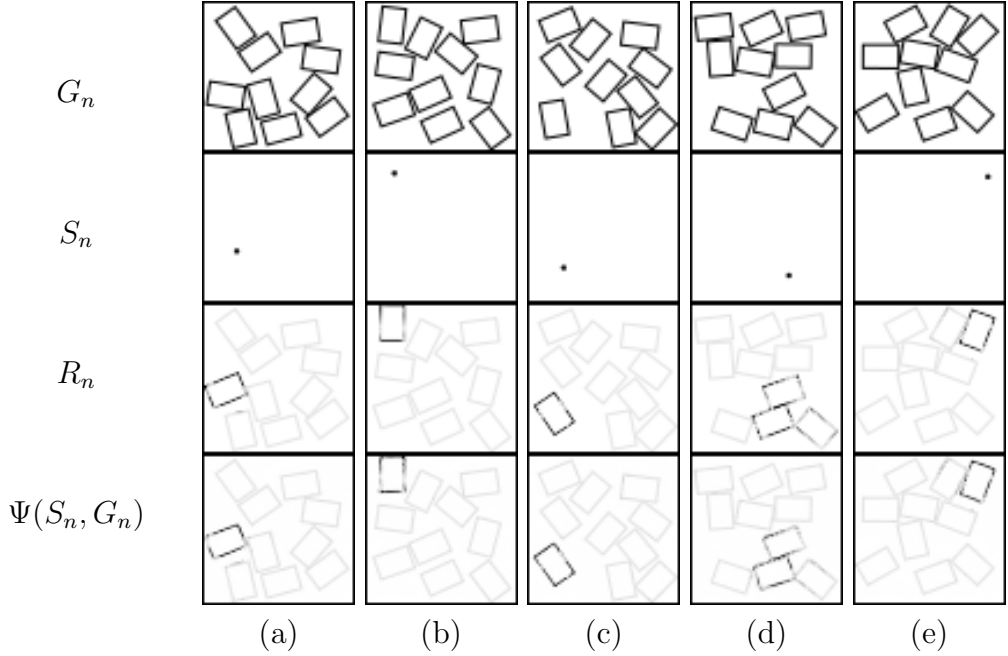
\begin{figure}[ht!]
\begin{center}
\hspace*{-0.75cm}
\raisebox{-3.3cm}{
\begin{tikzpicture}[scale=0.93]
\draw[draw=none] (0,0) -- (0.5,0) -- (0.5,8) -- (0,8) -- (0,0);
\node (n1) at (0.25, 7) {${{G_n}}$};
\node (n1) at (0.25, 5) {${{S_n}}$};
\node (n1) at (0.25, 3) {${{R_n}}$};
\node (n1) at (0.25, 1) {${\Psi({{S_n}}, {{G_n}})}$};
\end{tikzpicture}
}
\setlength{\tabcolsep}{2pt}
\renewcommand{\arraystretch}{1.0}
\begin{tabular}{ccccc}
\includegraphics[width=2cm]{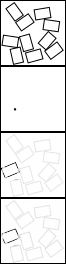} &
\includegraphics[width=2cm]{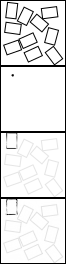} &
\includegraphics[width=2cm]{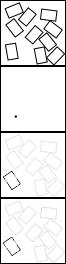} &
\includegraphics[width=2cm]{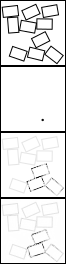} &
\includegraphics[width=2cm]{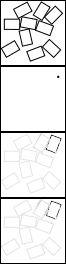} \\
(a) & (b) & (c) & (d) & (e)
\end{tabular}
\end{center}
\caption{Some illustrative prediction results. Each column corresponds
  to an example. The first row shows the starting configuration
  ${{S_n}}$, the second row the ``grasping point'' image ${{G_n}}$,
  the third row the true resulting configuration ${{R_n}}$, and the
  last row the predicted resulting configuration ${\Psi({{S_n}},
    {{G_n}})}$. {\bf For clarity, we highlight the pixels in the two
    bottom rows proportionally to the difference with the starting
    configuration.}
  See \S~\ref{sec:prediction} for discussion.\label{fig:results}}
\end{figure}

\begin{figure}[ht!]
\center
\setlength{\tabcolsep}{2pt}
\renewcommand{\arraystretch}{1.0}
\begin{tabular}{cccccc}
1/1024 &
2/1024 &
3/1024 &
4/1024 &
5/1024 &
6/1024 \\
(0.012389) &
(0.012250) &
(0.012055) &
(0.011718) &
(0.010772) &
(0.010062) \\
\includegraphics[width=2cm]{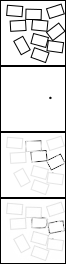} &
\includegraphics[width=2cm]{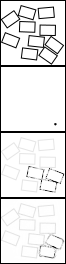} &
\includegraphics[width=2cm]{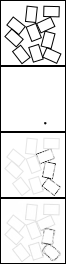} &
\includegraphics[width=2cm]{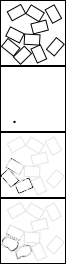} &
\includegraphics[width=2cm]{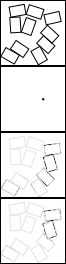} &
\includegraphics[width=2cm]{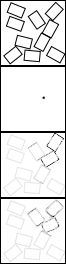} \\
(a) &
(b) &
(c) &
(d) &
(e) &
(f)
\end{tabular}

\vspace*{1em}

\setlength{\tabcolsep}{2pt}
\renewcommand{\arraystretch}{1.0}
\begin{tabular}{cccccc}
509/1024 &
510/1024 &
511/1024 &
512/1024 &
513/1024 &
514/1024 \\
(0.001378) &
(0.001373) &
(0.001373) &
(0.001360) &
(0.001356) &
(0.001355) \\
\includegraphics[width=2cm]{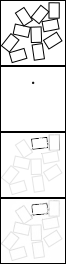} &
\includegraphics[width=2cm]{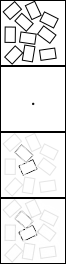} &
\includegraphics[width=2cm]{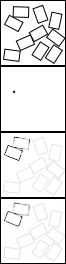} &
\includegraphics[width=2cm]{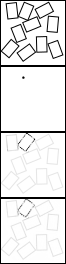} &
\includegraphics[width=2cm]{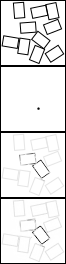} &
\includegraphics[width=2cm]{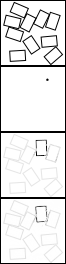} \\
(g) &
(h) &
(i) &
(j) &
(k) &
(l)
\end{tabular}
\caption{Examples from a random set of $1024$ test examples, selected
  according to their loss. Examples (a)-(f) are the worst regarding
  loss, (g)-(l) are median. The ranks (from worst to best) and $L^2$
  losses are provided above each example.  As in
  Figure~\ref{fig:results}, the four images in each example correspond
  from top to bottom to ${{F_n}}, {{G_n}}, {{S_n}}$ and $\Psi({{S_n}},
  {{G_n}})$.  {\bf For clarity, we highlight the pixels in the two
    bottom rows proportionally to the difference with the starting
    configuration.}  See \S~\ref{sec:prediction} for
  discussion.}\label{fig:results-ranked}
\end{figure}

The resulting network makes an accurate prediction of the final
configuration. We provide on Figure~\ref{fig:results} five examples
selected to illustrate the strengths and weaknesses of the prediction,
and on Figure~\ref{fig:results-ranked} some examples taken according
to the ranks of their individual losses to get a better intuition of
the overall performance.

We observe that the network:
\begin{itemize}
\item detects the grasped rectangle, and moves it while keeping the
  other ones undisturbed if there is no collision.
\item models translation and torque.
\item propagates to some extent the dynamics when collisions occur
  (Figure~\ref{fig:results}(d)).
\item models the hard borders around the area, although with some
  deformations (Figure~\ref{fig:results}(b)).
\item implements the synthesis of the perturbed scene, which involves
  in particular the segmentation of the moving vs. non-moving parts,
  and synthesis of edges at multiple orientations.
\end{itemize}

Figure~\ref{fig:results-ranked} gives a better understanding of the
overall performance of the network on $1,024$ test sequences. The six
examples (a)-(f) shown in the top half are those with the highest
losses, hence are the worst mistakes of the network, and the six
bottom examples (g)-(l) correspond to the ones ranked in the
middle.

It is remarkable that the worst mistakes correspond to complicated
cases even for a human, where the predicted motion would have been
correct for a minimal variation of the starting conditions (e.g.
Figure~\ref{fig:results-ranked}(a), and (b)), or involves a chain of
collisions (e.g. Figure~\ref{fig:results-ranked}(c), (d), (e), and
(f))



\subsection{Inner representation}\label{sec:inner-representation}

\begin{figure}[ht!]
\center
\includegraphics[width=15cm]{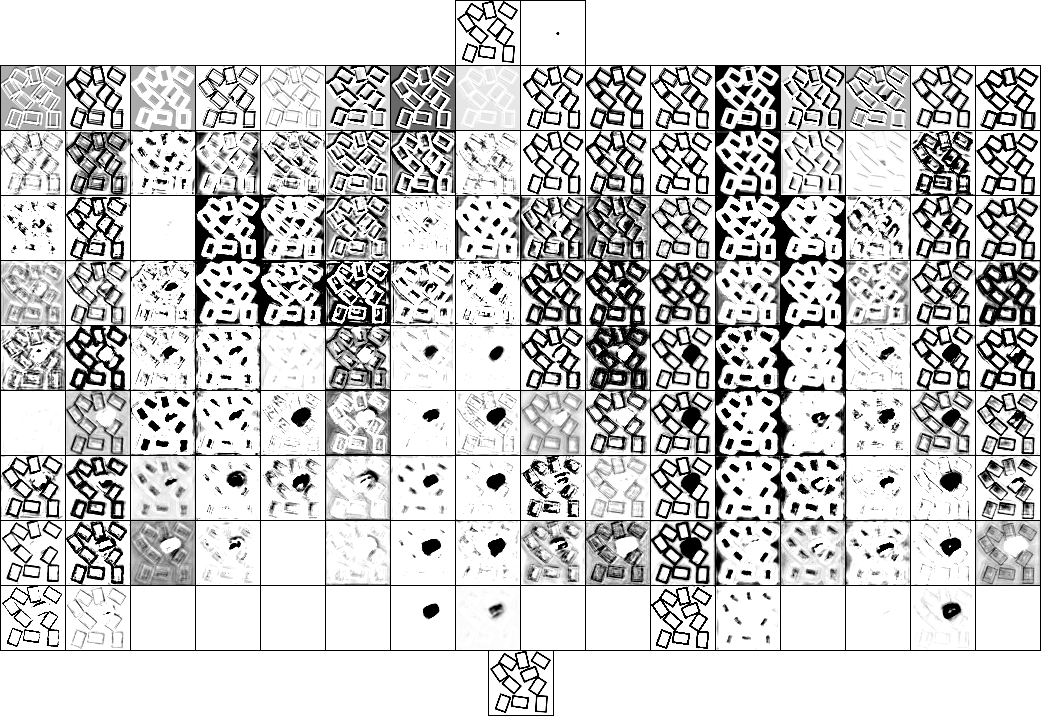}

\vspace*{0.5em}

(a)

\vspace*{1em}

\includegraphics[width=15cm]{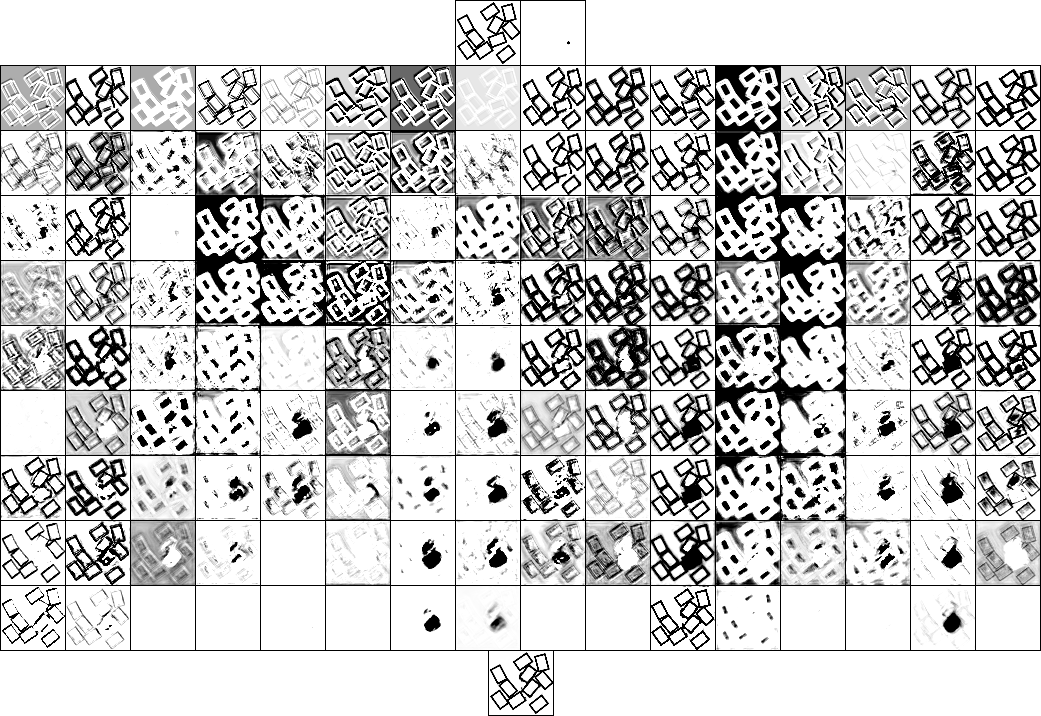}


\vspace*{0.5em}

(b)

\caption{Activations in the input, internal ReLU, and output
  layers. See \S~\ref{sec:inner-representation} for
  discussion.}\label{fig:internals}
\end{figure}

To shade a light on the processing occurring in the network, we
represent on Figure~\ref{fig:internals} for two examples the
processing from top to bottom as the activations of the input layer,
ReLU layers after each resnet module, and output layer.

The top row contains two activation maps corresponding to the two
input channels, respectively the starting configuration ${{S_n}}$ and
the grasping point ${{G_n}}$, the bottom row contains a single map,
corresponding to the network's output $\Psi({{S_n}}, {{G_n}})$. The
$9$ other rows correspond to the ReLU layer situated after the initial
convolution layer $\nnC^{f}_{{{2}}, {{q}}}$ that converts the $2$
input channels to the $q$ internal channels, followed by the ReLU
layers placed at the output of each of the ${{D}} = 8$ resnet modules
$\nnM$. The $16$ columns in this $9$ rows correspond to the ${{q}}=16$
channels for the internal coding.

We observe a homogeneity ``per channel'', which translates here to
``per column''. This is probably because the resnet architecture
favors processing near the identity, which results in gradual changes
through layers, and discourages the shuffling of information across
channels.

As we can see, an important part of the computation aims at segmenting
the grasped rectangle (channels $11$ and $15$), segmenting the moving
rectangles (channels $4$ and $5$), and removing the moving parts
(channels $1$, $10$, and $16$).

\clearpage

\appendix

\section{Torch network structure}\label{sec:torch-netw-struct}

{\small

\begin{verbatim}
nn.Sequential {
  [input -> (1) -> (2) -> (3) -> (4) -> (5) -> output]
  (1): nn.SpatialConvolution(2 -> 16, 5x5, 1,1, 2,2)
  (2): nn.SpatialBatchNormalization
  (3): nn.ReLU
  (4): nn.Sequential {
    [input -> (1) -> (2) -> (3) -> (4) -> (5) -> (6) -> (7) -> (8)
           -> (9) -> (10) -> (11) -> (12) -> (13) -> (14) -> (15) -> (16)
           -> (17) -> (18) -> (19) -> (20) -> (21) -> (22) -> (23) -> (24)
           -> (25) -> (26) -> (27) -> (28) -> (29) -> (30) -> (31) -> (32)
           -> output]
    (1): nn.ConcatTable {
      input
        |`-> (1): nn.Sequential {
        |      [input -> (1) -> (2) -> (3) -> (4) -> output]
        |      (1): nn.SpatialConvolution(16 -> 16, 5x5, 1,1, 2,2)
        |      (2): nn.SpatialBatchNormalization
        |      (3): nn.ReLU
        |      (4): nn.SpatialConvolution(16 -> 16, 5x5, 1,1, 2,2)
        |    }
         `-> (2): nn.Identity
         ... -> output
    }
    (2): nn.CAddTable
    (3): nn.SpatialBatchNormalization
    (4): nn.ReLU

    /... repeated 7 more times .../

  }
  (5): nn.SpatialConvolution(16 -> 1, 5x5, 1,1, 2,2)
}
\end{verbatim}
}


\clearpage

\bibliographystyle{abbrvnat}
\bibliography{dyncnn}

\typeout{get arXiv to do 4 passes: Label(s) may have changed. Rerun}

\end{document}